\documentclass[a4paper,12pt]{article}
\usepackage[french]{babel}
\usepackage[T1]{fontenc}
\usepackage[latin1]{inputenc}
\usepackage{epsfig}
\usepackage{amssymb}
\usepackage{amsmath}
\usepackage{tabls}

\newcommand{\conj}{\mathrm{Conj}}
\newcommand{\DP}{\mathrm{DP}}
\newcommand{\Y}{\mathrm{Y}}
\newcommand{\DS}{\mathrm{DS}}
\newcommand{\PCR}{\mathrm{PCR}}
\newcommand{\PCRmo}{\mathrm{PCR_{MO}}}
\newcommand{\PCRdsm}{\mathrm{PCR5}}
\newcommand{\Pmof}{\mathrm{PCR_{MOf}}}
\newcommand{\Pmog}{\mathrm{PCR_{MOg}}}

\newcommand{\bel}{\mathrm{bel}}
\newcommand{\pl}{\mathrm{pl}}
\newcommand{\betP}{\mathrm{betP}}

\newcommand{\eR}{\mbox{I\hspace{-.15em}R}}

\newcommand{\ind}{\mbox{1\hspace{-.25em}l}}

\title{Une nouvelle règle de combinaison répartissant le conflit - Applications en imagerie Sonar et classification de cibles Radar}

\author{Arnaud MARTIN et Christophe OSSWALD}

\date{}

\begin{document}
\maketitle

\paragraph{Résumé}
Le problème du conflit intrinsèque à la combinaison d'informations a poussé à de nombreuses réflexions ces dernières années, en particulier dans le cadre de la théorie des fonctions de croyance. Nous pouvons résumer les solutions apportées par trois façons de considérer le problème~: premièrement, nous pouvons chercher à réduire voire supprimer le conflit avant la combinaison d'informations, deuxièmement nous pouvons gérer le conflit de façon à ce qu'il n'intervienne pas lors de la combinaison et nous le considérons seulement lors de la décision finale, et troisièmement nous pouvons tenir compte du conflit lors de l'étape de combinaison. Si la première solution paraît la meilleure elle n'est pas toujours réalisable ou suffisante. Il peut être difficile de chercher à départager philosophiquement les deux dernières stratégies. Cependant d'un point de vue applicatif seule la décision compte, et c'est donc dans cette optique que nous chercherons à les comparer. 
Nous proposons ici une nouvelle règle qui a pour principe de répartir le conflit proportionnellement sur les éléments produisant ce conflit. Nous comparons les différentes règles à partir de données réelles en imagerie Sonar et en classification de cibles Radar.

{\bf Mots Clés~: Fonctions de croyance, Conflit, Règle de Combinaison, Imagerie Sonar, Classification de cibles Radar}

\paragraph{Abstract}
These last years, there were many studies on the problem of the conflict coming from information combination, especially in evidence theory. We can summarise the solutions for manage the conflict into three different approaches: first, we can try to suppress or reduce the conflict before the combination step, secondly, we can manage the conflict in order to give no influence of the conflict in the combination step, and then take into account the conflict in the decision step, thirdly, we can take into account the conflict in the combination step. The first approach is certainly the better, but not always feasible. It is difficult to say which approach is the best between the second and the third. However, the most important is the produced results in applications. We propose here a new combination rule that distributes the conflict proportionally on the element given this conflict. We compare these different combination rules on real data in Sonar imagery and Radar target classification.

{\bf Key Words~: Belief Functions, Conflict, Combination Rule, Sonar Imagery, Radar Target Classification}

\section{Introduction}
La fusion d'informations a depuis plusieurs années permis d'apporter des solutions à la combinaison d'informations issues de diverses sources afin d'améliorer la prise de décision par exemple dans un système de classification. Le fait de devoir tenir compte de plusieurs sources a poussé le développement de plusieurs théories permettant de modéliser finement les informations issues de ces sources en terme de fiabilité, incertitude, imprécision ou encore incomplétude. Parmi ces théories de l'incertain, la théorie des fonctions de croyance issue des recherches de Dempster et Shafer \cite{Dempster67,Shafer76} a connu ces dernières années un développement important notamment dans sa mise en forme pour de nombreuses applications. 

La théorie des fonctions de croyance est fondée sur la manipulation des fonctions de masse (ou masse élémentaire de croyance). Les fonctions de masse sont définies sur l'ensemble de toutes les disjonctions du cadre de discernement $\Theta=\{C_1,\ldots,C_n\}$ et à valeurs dans $[0,1]$, où les $C_i$ représentent les hypothèses supposées exhaustives et exclusives. Cet ensemble est noté $2^\Theta$. Généralement, il est ajouté une condition de normalité, donnée par :
\begin{eqnarray}
\label{hyp1_fonction_masse}
\sum_{X \in 2^\Theta} m_j(X)=1,
\end{eqnarray}
où $m_j(.)$ représente la fonction de masse pour une source (ou un expert) $S_j$, $j=1,...,M$. Les éléments $X$ tels que $m(X)>0$ sont appelés les éléments focaux. La première difficulté est donc de définir ces fonctions de masse selon le problème. A partir de ces fonctions de masse, d'autres fonctions de croyance peuvent être définies, telles que les fonctions de crédibilité, représentant l'intensité que toutes les sources croient que les éléments focaux d'une source affirment la croyance en un élément. Elles sont données pour tout $X \in 2^\Theta$ par~:
\begin{eqnarray}
\bel(X)=\sum_{Y \in 2^X, Y \neq \emptyset} m(Y),
\end{eqnarray}
ou encore les fonctions de plausibilité, représentant l'intensité avec laquelle on ne doute pas en un élément, données pour tout $X \in 2^\Theta$ par~:
\begin{eqnarray}
\label{pignistic}
\pl(X)=\sum_{Y \in 2^\Theta, Y\cap X \neq \emptyset} m(Y)=\bel(\Theta)-\bel(X^c)=1-m(\emptyset)-\bel(X^c),
\end{eqnarray}
où $X^c$ est le complémentaire de $X$. 

Afin de conserver un maximum d'informations, il est préférable de rester à un niveau crédal ({\em i.e.} de manipuler des fonctions de croyance) pendant l'étape de combinaison des informations pour prendre la décision sur les fonctions de croyance issues de la combinaison. Si la décision prise par le maximum de crédibilité peut être trop pessimiste, la décision issue du maximum de plausibilité est bien souvent trop optimiste. Le maximum de la probabilité pignistique, introduite par \cite{Smets90b}, reste le compromis le plus employé. La probabilité pignistique est donnée pour tout $X \in 2^\Theta$, avec $X \neq \emptyset$ par~:
\begin{eqnarray}
\betP(X)=\sum_{Y \in 2^\Theta, Y \neq \emptyset} \frac{|X \cap Y|}{|Y|} \frac{m(Y)}{1-m(\emptyset)}.
\end{eqnarray}
Notons que nous obtenons ainsi une probabilité peu conforme à la notion de fonction de masse. Le maximum de probabilité pignistique est généralement considéré uniquement sur les hypothèses singletons $C_i$ à cause de l'additivité des probabilités. En effet~: 
$$\betP(C_1 \cup C_2) \geq \betP(C_i), \,\, i=1,2.$$
De la même façon la croissance des fonctions de crédibilité et de plausibilité pousse à prendre une décision uniquement sur les singletons. Si l'application autorise à prendre une décision finale sur une disjonction, il est alors préférable de considérer les fonctions de masse directement ou de développer une autre fonction de croyance non croissante adaptée à l'application. Dans un cas général une fonction de croyance non croissante sur le treillis d'inclusion mène à des résultats rapidement contre intuitif. Il peut également être intéressant de ne plus considérer l'unique hypothèse la plus vraisemblable, mais un sous-ensembles des hypothèses les plus vraisemblables \cite{Appriou05}.

Chercher à combiner des informations issues de plusieurs sources qui ne sont pas en accord fait bien souvent apparaître un conflit entre les sources. Ce conflit peut provenir d'un manque d'exhaustivité des sources, ou encore d'un manque de fiabilité de celles-ci \cite{Appriou02}. Plusieurs approches sont envisageables pour gérer ce conflit. Dans le cas d'un manque d'exhaustivité des sources, une approche classique est la technique du {\it hedging} qui consiste à ajouter un élément au cadre de discernement. Lorsque la fiabilité des sources peut être estimée, une approche consiste à affaiblir les masses selon cette fiabilité. Cependant, depuis le problème posé par Zadeh, que nous reprendrons plus loin, de nombreuses règles de combinaison ont été proposées afin de manager le conflit \cite{Yager87, Dubois88, Smets90, Inagaki91, Smets97, Lefevre02a, Josang03, Daniel04, Smarandache05, Florea06}.

Nous proposons ici une nouvelle règle qui a pour principe de répartir le conflit proportionnellement sur les éléments produisant ce conflit. Dans un premier temps nous rappelons les différentes règles de combinaison, puis nous montrons en quoi il peut être intéressant de considérer cette nouvelle règle. Les différentes règles proposées au cours de ces dernières années peuvent être performantes selon les applications considérées. Dans une dernière partie, nous étudions le comportement de ces règles en terme de décision, dans le cadre de deux applications à partir de données réelles~: en imagerie Sonar et en reconnaissance de cibles Radar.

\section{La combinaison}

Différentes approches de combinaison des fonctions de masse ont été proposées. La règle orthogonale de Dempster-Shafer non normalisée proposée par Smets \cite{Smets90}, est définie pour deux fonctions de masse $m_1$ et $m_2$ et pour tout $A \in 2^\Theta$ par :
\begin{eqnarray}
m_\conj(A)=\displaystyle \sum_{B\cap C =A} m_1(B)m_2(C):=(m_1\oplus m_2 )(A).
\end{eqnarray}
Elle est donnée pour $M$ experts, pour tout $X \in 2^\Theta$ par~:
\begin{eqnarray}
\label{conjunctive}
m_\conj(X)=\sum_{Y_1 \cap ... \cap Y_M = X} \prod_{j=1}^M m_j(Y_j):=(\mathop{\oplus}_{j=1}^M m_j)(A),
\end{eqnarray}
où $Y_j \in 2^\Theta$ est un élément focal de la fonction de masse de l'expert $j$, et $m_j(Y_j)$ la masse associée.

Cette opérateur est associatif et commutatif. La masse affectée sur l'ensemble vide s'interprète comme une mesure de conflit. Ce conflit peut provenir d'un manque d'exhaustivité des sources, ou encore d'un manque de fiabilité \cite{Appriou02}. Dans le premier cas le fait d'avoir une masse non nulle sur l'ensemble vide (cas d'un monde ouvert) est concevable. Si l'on souhaite rester en monde fermé, la technique du {\it hedging} consiste à répartir le conflit sur le nouvel élément $e$, et la règle est alors donnée par~:
\begin{eqnarray}
\label{hedging}
\begin{array}{rcl}
m_h(X)&=&m_\conj(X), \, \forall X\neq \emptyset \\
m_h(e)&=&m_\conj(\emptyset).
\end{array}
\end{eqnarray}

Si les sources ne sont pas fiables, lorsqu'il est possible de quantifier la fiabilité de chacune des sources il est important de procéder à un affaiblissement en redéfinissant les fonctions de masse par~:
\begin{eqnarray}
\left\{
\begin{array}{l}
	m'_j(X)=\alpha_j m_j(X), \, \forall X \in 2^\Theta \\
	m'_j(\Theta)=1-\alpha_j (1-m_j(\Theta)).
\end{array}
\right.
\end{eqnarray}
$\alpha_j \in [0,1]$ est le coefficient d'affaiblissement de la source $S_j$ qui est alors une estimation de la fiabilité de l'expert $S_j$, éventuellement comme une fonction de $X \in 2^\Theta$.

Les experts pouvant s'exprimer dans $2^\Theta$ (s'ils ne sont ni sûrs ni précis), l'apparition de conflit est inévitable. En effet, la règle orthogonale n'est pas idempotente, ainsi nous pouvons définir un \textit{auto-conflit} d'ordre $n$ pour chaque source $j$ par~: 
\begin{eqnarray}
\label{autoconf}
a_n(j)=\displaystyle \mathop{\oplus}_{k=1}^n m_j(\emptyset),
\end{eqnarray}
où l'opérateur $\mathop{\oplus}$ est l'opérateur de combinaison conjonctive de l'équation (\ref{conjunctive}) et nous avons la propriété~:
\begin{eqnarray}
a_n(j)\leq a_{n+1}(j).
\end{eqnarray}

Initialement, Dempster et Shafer ont proposé une règle orthogonale normalisée, afin de rester en monde fermé. Ainsi la répartition du conflit se fait de manière uniforme lors de la combinaison, pour tout $X \in 2^\Theta$, $X\neq \emptyset$  par~:
\begin{eqnarray}
\label{DS}
m_\DS(X)=\frac{1}{1-m_\conj(\emptyset)} \sum_{Y_1 \cap ... \cap Y_M = X}  \prod_{j=1}^M m_j(Y_j)=\frac{m_\conj(X)}{1-m_\conj(\emptyset)},
\end{eqnarray}
où $Y_j \in 2^\Theta$ est la réponse de l'expert $j$, et $m_j(Y_j)$ la fonction de masse associée.

Smets ne répartit le conflit établi sur l'ensemble vide lors de la combinaison, qu'à l'étape de décision en prenant le maximum de la probabilité pignistique. Il multiplie toutes les masses par $\displaystyle \frac{1}{1-m_\conj(\emptyset)}$ ({\em cf.} équation (\ref{pignistic})). Ce critère de décision présente un compromis entre une décision pessimiste par le maximum de crédibilité et une décision optimiste par le maximum de plausibilité, aussi bien en monde ouvert qu'en monde fermé. Mais ces trois critères produisent la même décision que l'on normalise lors de l'étape de combinaison ou bien lors de la décision. 

La gestion du conflit peut cependant être réalisée lors de la combinaison de manière différente que dans le cas de la règle orthogonale normalisée, comme nous le verrons en rappelant l'exemple de Zadeh. Ainsi Yager \cite{Yager87} répartit le conflit sur l'ignorance totale ({\em i.e.} sur la masse de $\Theta$) afin de rester en monde fermé et considère qu'on ne sait rien en cas de conflit. Ainsi la règle qu'il propose est donnée par~:
\begin{eqnarray}
\label{Yager}
\begin{array} {rcl}
m_\Y(X)&=& m_\conj(X), \forall X \in 2^\Theta, \, X\neq \emptyset, \, X \neq \Theta \\
m_\Y(\Theta)&=&m_\conj(\Theta)+m_\conj(\emptyset)\\
m_\Y(\emptyset)&=&0.
\end{array}
\end{eqnarray}

Dans la plupart des applications, nous cherchons à prendre une décision sur les singletons et non sur l'ensemble $2^\Theta$. C'est ce qui est réalisé par la décision du maximum de probabilité pignistique. Ainsi, le problème soulevé par Zadeh ({\em cf.} tableau \ref{ex_zadeh}) lors d'un fort conflit n'est pas résolu, et la décision sur les singletons reste $C$.

\begin{table}[htbp]
\caption{Exemple de Zadeh}
\begin{center}
\begin{tabular}{|c|c|c|c|c|c|c|} \hline
        & $\emptyset$ & $A$ & $B$ & $A\cup B$ & $C$ & $\Theta$  \\ \hline
       $m_1$ & 0 & 0.9 & 0 & 0 & 0.1 &0  \\ \hline
       $m_2$ & 0 & 0 & 0.9 & 0 & 0.1 &0    \\ \hline
        $m_\DS$ & 0 & 0 & 0& 0& 1 & 0 \\ \hline
       $m_\conj$ & 0.99 & 0 & 0& 0& 0.01 & 0 \\ \hline
       $m_\Y$ & 0& 0 & 0& 0& 0.01  & 0.99  \\ \hline
\end{tabular}
\end{center}
\label{ex_zadeh}
\end{table}

Dubois et Prade \cite{Dubois88} ont proposé une gestion plus fine du conflit en répartissant le conflit partiel (par exemple issu uniquement de deux sources, l'une annonçant $A$ et l'autre $B$) sur les ignorances partielles (c'est-à-dire $A\cup B$). Cette règle est donnée pour tout $X \in 2^\Theta$, $X\neq \emptyset$ par~: 
\begin{eqnarray}
\label{DP}
m_\DP(X)=\sum_{Y_1 \cap ... \cap Y_M = X} \prod_{j=1}^M m_j(Y_j)+\sum_{
\scriptsize \begin{array}{c}
Y_1 \cup ... \cup Y_M = X\\
Y_1 \cap ... \cap Y_M = \emptyset \\
\end{array}} \prod_{j=1}^M m_j(Y_j),
\end{eqnarray}
où $Y_j \in 2^\Theta$ est un élément focal de l'expert $j$, et $m_j(Y_j)$ la fonction de masse associée. Si nous reprenons l'exemple de Zadeh, nous obtenons~:
 \\
\begin{center}
  \begin{tabular}{|l|c|c|c|c|c|c|c|c|c|}
    \hline
    & $\emptyset$ & $A$ & $B$ & $A\cup B$ & $C$& $A\cup C$ & $B\cup C$ & $\Theta$ \\
    \hline
    $m_\DP$ & 0 & 0& 0& 0.99 & 0.01 & 0 & 0 & 0 \\
    \hline
    $\bel_\DP$  &0     &    0     &    0   & 0.99 &   0.01  &  0.01  &  0.01  &  1\\
    \hline
    $\pl_\DP$  &0  &  0.99  &  0.99  &  0.99 &   0.01  &  1 & 1 & 1 \\
    \hline
    $\betP_\DP$  & 0 & 0.495  &  0.495  & 0.99 &  0.01 & 0.505 & 0.505 & 1 \\
    \hline
  \end{tabular}
\end{center}
$$ $$
Ainsi avec le maximum de probabilité pignistique ou de la plausibilité nous choisissons $A$ ou $B$ sans pouvoir les distinguer à cause de l'égalité des masses données par les différents experts sur $A$ et $B$. Si ces masses sont différentes la décision est alors possible, comme l'illustre l'exemple choisi suivant~:
\\
\begin{center}
  \begin{tabular}{|l|c|c|c|c|c|c|c|c|c|}
    \hline
    & $\emptyset$ & $A$ & $B$ & $A\cup B$ & $C$& $A\cup C$ & $B\cup C$ & $\Theta$ \\
    \hline
    $m_1$ & 0 & 0.5421& 0.0924& 0 & 0.2953 & 0 & 0 & 0.0702 \\
    \hline
    $m_2$ & 0 & 0.2022& 0.0084& 0 & 0.6891 & 0 & 0 & 0.1003 \\
    \hline
    $m_\DP$ & 0 & 0.1782 & 0.0106 & 0.0233 & 0.2815 & 0.4333 & 0.0662 & 0.007 \\
    \hline
    $\bel_\DP$  & 0 & 0.1782  &  0.0106   & 0.2120  &  0.2815  &  0.8929  &  0.3583 & 1\\
    \hline
    $\pl_\DP$  &0  &  0.6417  &  0.1071  &  0.7185  &  0.7880  &  0.9894   & 0.8218 & 1 \\
    \hline
    $\betP_\DP$  & 0 & 0.4088  &  0.0577  & 0.4665 &  0.5335 & 0.9423 &0.5912 & 1 \\
    \hline
  \end{tabular}
\end{center}
$$ $$
Sur cet exemple la décision prise sera donc la classe $C$.

De manière générale, la répartition du conflit peut s'écrire \cite{Inagaki91,Lefevre02a,Lefevre02b}~:
\begin{eqnarray}
\label{rep}
m_c(X)=m_\conj(X)+w(X)m_\conj(\emptyset),
\end{eqnarray}
où $\displaystyle \sum_{X\in 2^\Theta} w(X)=1$. Toutes les règles précédentes peuvent être vues comme un cas particulier de celle-ci, la difficulté étant le choix de poids $w(X)$. De plus cette règle ne fait pas apparaître la gestion fine des conflits partiels telle quelle est considérée dans la règle de Dubois et Prade, même s'il est toujours possible d'en tenir compte selon la définition des poids $w(X)$.

A notre connaissance les seules règles répartissant les conflits partiels sur les éléments dont la combinaison crée le conflit, sont la règle du $minC$ proposée par \cite{Daniel04} et celles proposées par Dezert et Smarandache \cite{Smarandache05, Florea06}. Ces règles peuvent également être vues comme des cas particuliers de la répartition du conflit total de l'équation (\ref{rep}) avec des poids choisis de manière adéquate. La règle la plus aboutie de Dezert et Smarandache est la PCR5 donnée pour 2 experts par~:
\begin{eqnarray}
\label{PCR5_2}
\begin{array}{l}
m_\PCRdsm(X)=m_{12}(X)+\\
\displaystyle
\sum_{\begin{array}{l}
\scriptstyle Y\in 2^\Theta, \\
\scriptstyle X\cap Y=\emptyset
\end{array}} \left(\frac{m_1(X)^2 m_2(Y)}{m_1(X)+m_2(Y)}+\frac{m_2(X)^2 m_1(Y)}{m_2(X)+m_1(Y)}\right),
\end{array}
\end{eqnarray}
où $m_{12}(.)$ est la combinaison orthogonale pour deux experts, et le dénominateur est non nul.

L'extension à $M$ experts peut s'écrire~:

\begin{eqnarray}
\label{PCR5combination}
\begin{array}{rcl}
  \displaystyle \!\!\!m_\PCRdsm(X) &=& m_\conj(X) \\ 
  
   \!\!\!\!\!\!\!\!\!\!\!\!\!&+&\displaystyle \sum_{i=1}^M
  m_i(X)  \!\!\!\!\!\!\!\!\!\!\! \sum_{\begin{array}{c}
      \scriptstyle {\displaystyle \mathop{\cap}_{\scriptscriptstyle
	  k=1}^{\scriptscriptstyle M\!-\!1}} Y_{\sigma_i(k)} \cap X = \emptyset \\
      \scriptstyle (Y_{\sigma_i(1)},...,Y_{\sigma_i(M\!-\!1)})\in (2^\Theta)^{M\!-\!1}
  \end{array}}  \!\!
  \frac{
    \displaystyle \Bigg(\prod_{j=1}^{M\!-\!1} m_{\sigma_i(j)}(Y_{\sigma_i(j)})\ind_{j>i}\Bigg)
    \!\!\!\prod_{Y_{\sigma_i(j)}=X} \!\!\!\!\! m_{\sigma_i(j)}(Y_{\sigma_i(j)})
  }{
    \displaystyle \!\!\!\!\!\!\!\!\!\!\!\!\sum_{~~~~~\renewcommand{\arraystretch}{1.8}\begin{array}{c}
	\scriptstyle Z\in\{X, Y_{\sigma_i(1)}, \ldots, Y_{\sigma_i(M\!-\!1)}\}
    \end{array}}
    \!\!\!\!\!\!\!\!\!\!\!\!\!\!\!\!\!\!\!\!\!\!\!\!\!\prod_{Y_{\sigma_i(j)}=Z}^{\:}\!\!\!\!\!\!\!
    \big(m_{\sigma_i(j)}(Y_{\sigma_i(j)}).T({\scriptstyle X\!=\!Z,m_i(X)})\big)},
\end{array}
\end{eqnarray}
où $\sigma_i$ prend des valeurs de 1 à $M$ selon $i$~:
\begin{eqnarray}
\label{sigma}
\left\{
\begin{array}{ll}
\sigma_i(j)=j &\mbox{si~} j<i,\\
\sigma_i(j)=j+1 &\mbox{si~} j\geq i,\\
\end{array}
\right.
\end{eqnarray}
et~:
\begin{eqnarray}
\label{T}
\left\{
\begin{array}{ll}
T(B,x)=x &\mbox{si $B$ est vraie},\\
T(B,x)=1 &\mbox{si $B$ est fausse}.\\
\end{array}
\right.
\end{eqnarray}

Si nous reprenons l'exemple de Zadeh, cette règle donne~:
$$m_\PCRdsm(A) = 0.486, ~~ m_\PCRdsm(B) = 0.486,~~ m_\PCRdsm(C) = 0.028.$$
La masse équivalente sur $A$ et $B$ provient des masses initiales des deux experts, pour lever l'indétermination il suffit d'une légère différence entre les deux experts. 

\section{Une nouvelle règle de combinaison répartissant le conflit}

La règle précédente (équation (\ref{PCR5combination})) n'est pas toujours satisfaisante comme nous le montrons dans les considérations ci-dessous. Ainsi nous avons proposé une autre extension à $M$ experts (équivalente dans le cas de deux experts)~:

\begin{eqnarray}
\label{PCR6combination}
\begin{array}{rcl}
  \displaystyle m_\PCRmo(X)  &=&  \displaystyle m_\conj(X) \\
  \!\!\!\!\!\!\!\!\!\!\!\!\!&+& \displaystyle \sum_{i=1}^M   m_i(X)^2 
  \!\!\!\!\!\!\!\! \displaystyle \sum_{\begin{array}{c}
      \scriptstyle {\displaystyle \mathop{\cap}_{k=1}^{M\!-\!1}} Y_{\sigma_i(k)} \cap X = \emptyset \\
      \scriptstyle (Y_{\sigma_i(1)},...,Y_{\sigma_i(M\!-\!1)})\in (2^\Theta)^{M\!-\!1}
  \end{array}}
  \!\!\!\!\!\!\!\!\!\!\!\!
  \left(\!\!\frac{\displaystyle \prod_{j=1}^{M\!-\!1} m_{\sigma_i(j)}(Y_{\sigma_i(j)})}
       {\displaystyle m_i(X) \!+\! \sum_{j=1}^{M\!-\!1} m_{\sigma_i(j)}(Y_{\sigma_i(j)})}\!\!\right)\!\!,
\end{array}
\end{eqnarray}
où $\sigma$ est défini par l'équation (\ref{sigma}).

Nous pouvons proposer deux règles encore plus générales données par~:
\begin{eqnarray}
\label{GenePCR_f}
\begin{array}{l}
m_\Pmof(X)=m_c(X)+\\
\displaystyle
\sum_{i=1}^M m_i(X)f(m_i(X)). 
\displaystyle
~~~~~~ \sum_{\begin{array}{c}
\scriptstyle (Y_1,...,Y_{M-1})\in (2^\Theta)^{M-1} \backslash \{X^{M-1}\}\\
\scriptstyle \cap_{k=1}^{M-1} Y_k \cap X = \emptyset\\
\end{array}} \frac{\displaystyle \prod_{j=1}^{M-1} m_{\sigma_i(j)}(Y_{\sigma_i(j)})}{f(m_i(X))+\displaystyle \sum_{j=1}^{M-1} f(m_{\sigma_i(j)}(Y_{\sigma_i(j)}))},
\end{array}
\end{eqnarray}
avec les mêmes notations que dans l'équation (\ref{PCR6combination}), et $f$ est une fonction croissante définie sur $]0,1]$ et à valeurs dans $\eR^{+*}$. 

\begin{eqnarray}
\label{GenePCR_g}
\begin{array}{l}
m_\Pmog(X)=m_c(X)+\\
\displaystyle
\sum_{i=1}^M m_i(X)g(m_i(X)).  
\displaystyle
~~~~~~ \sum_{\begin{array}{c}
\scriptstyle (Y_1,...,Y_{M-1})\in (2^\Theta)^{M-1} \backslash \{X^{M-1}\}\\
\scriptstyle \cap_{k=1}^{M-1} Y_k \cap X = \emptyset\\
\end{array}} \frac{\displaystyle \prod_{j=1}^{M-1} m_{\sigma_i(j)}(Y_{\sigma_i(j)})}{g\left(m_i(X)+\displaystyle \sum_{j=1}^{M-1} m_{\sigma_i(j)}(Y_{\sigma_i(j)})\right)},
\end{array}
\end{eqnarray}
avec les mêmes notations que dans l'équation (\ref{PCR6combination}), et $g$ est une fonction croissante définie sur $]0,1]$ et à valeurs dans $\eR^{+*}$. 

Par exemple, $f(x)=g(x)=x^\alpha$, avec $\alpha \in \eR^+$. Sur ces deux dernières règles, la difficulté du choix de la fonction $f$ ou $g$ s'apparente à celle du choix des poids dans l'équation (\ref{rep}).

\subsection{Remarques sur cette nouvelle règle de combinaison}
\begin{itemize}
\item Le problème principal des règles de redistribution du conflit est peut-être la non-associativité. Prenons l'exemple de trois experts et de deux classes donné par~:
\\
\begin{center}
  \begin{tabular}{|l|c|c|c|c|}
    \hline
    & $\emptyset$ & $A$ & $B$ & $\Theta$ \\
    \hline
    Expert 1 & 0 & 1 & 0 & 0 \\
    \hline
    Expert 2 & 0 & 0 & 1 & 0\\
    \hline
    Expert 3 & 0 & 0 & 1 & 0 \\
    \hline
  \end{tabular}
\end{center}
$$ $$
En fusionnant l'expert 1 et 2 puis l'expert 3, la règle donne~:
$$m_{12}(A) = 0.5, ~~ m_{12}(B) = 0.5,$$
et
$$m_{(12)3}(A) = 0.25, ~~ m_{(12)3}(B) = 0.75.$$

Maintenant, si nous fusionnons l'expert 2 et 3 puis l'expert 1, la règle donne~:
$$m_{23}(A) = 0, ~~ m_{23}(B) = 1,$$
et
$$m_{(12)3}(A) = 0.5, ~~ m_{(12)3}(B) = 0.5.$$
Ainsi le résultat est bien différent.

De plus avec l'équation (\ref{PCR5combination}) nous obtenons~:
$$m_{(123)}(A) = 1/2, ~~ m_{(123)}(B) = 1/2,$$
et avec la règle $\PCRmo$ nous obtenons~:
$$m_{(123)}(A) = 1/3, ~~ m_{(123)}(B) = 2/3,$$
qui est un résultat plus intuitif.

L'associativité peut être importante en fusion dynamique. Cependant, lors de la fusion dynamique il est courant d'affaiblir les masses afin de tenir compte plus faiblement des avis des experts s'étant exprimé il y a longtemps. La $\PCRmo$ le fait naturellement, certes sans avoir le choix des poids d'affaiblissement, mais il est aussi recommandable dans la mesure du possible de procéder à un affaiblissement avant la combinaison. 

\item Le conflit n'est pas seulement redistribué sur les singletons. En effet, prenons par exemple trois experts~:
\\
\begin{center}
  \begin{tabular}{|l|c|c|c|c|}
    \hline
    & $A\cup B$ & $B\cup C$ & $A\cup C$ & $\Theta$ \\
    \hline
    Expert 1  & 0.7 & 0 & 0 & 0.3 \\
    \hline
    Expert 2 & 0 & 0 & 0.6 & 0.4 \\
    \hline
    Expert 3 & 0 & 0.5 & 0 & 0.5 \\
    \hline
  \end{tabular}
\end{center}
$$ $$
Le conflit est ici donné par 0.7$\times$0.6$\times$0.5=0.21, avec la règle $\PCRmo$ nous obtenons~:
$$m_{(123)}(A) = 0.21,$$
$$m_{(123)}(B) = 0.14,$$
$$m_{(123)}(C) = 0.09,$$
$$m_{(123)}(A\cup B) = 0.14+0.21.\frac{7}{18}\simeq 0.222,$$
$$m_{(123)}(B \cup C) = 0.06+0.21.\frac{6}{18} = 0.13,$$
$$m_{(123)}(A \cup C) = 0.09+0.21.\frac{5}{18} \simeq 0.147,$$
$$m_{(123)}(\Theta) = 0.06.$$
Nous obtenons ainsi la classe $A$ quelque soit le critère de décision~:
\\
\begin{center}
  \begin{tabular}{|l|c|c|c|c|c|c|c|c|c|}
    \hline
    & $\emptyset$ & $A$ & $B$ & $A\cup B$ & $C$& $A\cup C$ & $B\cup C$ & $\Theta$ \\
    \hline
    $m_{(123)}$ & 0 & 0.21 & 0.14 & 0.2217 & 0.09 & 0.1483 & 0.13 & 0.06 \\
    \hline
    $\bel_{(123)}$  &0 & 0.21 & 0.14 & 0.5717 & 0.09 & 0.4483 & 0.36 & 1 \\
    \hline
    $\pl_{(123)}$  &0 & 0.64 & 0.5517 & 0.91 & 0.4283 & 0.86 & 0.79 & 1 \\
    \hline
    $\betP_{(123)}$  & 0 & 0.415 & 0.3358 &  0.7508 & 0.2492 & 0.6642 & 0.585 & 1 \\
    \hline
  \end{tabular}
\end{center}
$$ $$

\item La $\PCRmo$ apporte des décisions qui peuvent être différentes des autres règles de combinaison répartissant le conflit. Reprenons l'exemple choisi précédent~: 
\\
\begin{center}
  \begin{tabular}{|l|c|c|c|c|c|c|c|c|c|}
    \hline
    & $\emptyset$ & $A$ & $B$ & $A\cup B$ & $C$& $A\cup C$ & $B\cup C$ & $\Theta$ \\
    \hline
    $m_1$ & 0 & 0.5421& 0.0924& 0 & 0.2953 & 0 & 0 & 0.0702 \\
    \hline
    $m_2$ & 0 & 0.2022& 0.0084& 0 & 0.6891 & 0 & 0 & 0.1003 \\
    \hline
    $m_\PCRmo$ & 0 & 0.5421 & 0.0924  & 0  &  0.2953 & 0 & 0& 0.0702 \\
    \hline
    $\bel_\PCRmo$  & 0 & 0.5421  &  0.0924   & 0.6345 &   0.2953  &  0.8374  &  0.3877 & 1\\
    \hline
    $\pl_\PCRmo$  &0  &  0.6123  &  0.1626  & 0.7047  &  0.3655  &  0.9076 &   0.4579 & 1 \\
    \hline
    $\betP_\PCRmo$  & 0 & 0.5655  &  0.1158  & 0.6813 &  0.3187 & 0.8842 & 0.4345 & 1 \\
    \hline
  \end{tabular}
\end{center}
$$ $$
Sur cet exemple la classe retenue par la $\PCRmo$ sera donc la classe $A$, contrairement à la règle de combinaison de Dubois et Prade (équation (\ref{DP}) qui donne la classe $C$. En dehors des considérations philosophiques sur l'intérêt de reporter les conflits partiels sur les ignorances partielles puis sur les singletons lors de l'étape de décision ou l'intérêt de reporter ceux-ci directement sur les singletons lors de l'étape de combinaison, ce qui importe est la règle donnant les meilleures performances. L'exemple précédent montre qu'il existe des situations dans lesquelles la décision est différente sans pour autant qu'il soit possible de dire quelle règle est la meilleure. Pour ce faire nous allons étudier le comportement de ces règles dans le cadre de deux applications où la réalité est supposée connue dans la section \ref{appli}.

\end{itemize}

\subsection{La combinaison des fonctions de croyance à valeurs réelles en répartissant le conflit}
L'article récent de Ph. Smets \cite{Smets05} propose l'extension des fonctions de croyance sur les nombres réels. Cette extension ouvre la porte à de nombreuses applications des fonctions de croyance dans le domaine continu telles que l'estimation ou le suivi de cibles. Les fonctions de masse à valeurs réelles sont définies sur des intervalles, par exemple de $\eR$. Pour les mêmes raisons que dans le domaine discret, les experts donnant une masse sur des intervalles disjoints entrent en conflit. La combinaison de Smets \cite{Smets05} pour deux experts s'exprimant sur $[a_1,b_1]$ et sur $[a_2,b_2]$, dont l'intersection non vide est donnée par $[a_1,b_1]\cap [a_2,b_2]=[\max(a_1,a_2),\min(b_1,b_2)]=[a,b]$ est donnée par~:

\begin{eqnarray}
\label{conj_cont}
\begin{array}{l}
m_\conj([a,b])=\displaystyle \int_{x=-\infty}^a \int_{y=b}^{+\infty} m_1([x,b])m_2([a,y]) dy dx \\
+\displaystyle \int_{x=-\infty}^a \int_{y=b}^{+\infty}m_1([a,y]) m_2([x,b]) dy dx  \\
+\displaystyle m_1([a,b])\int_{x=-\infty}^a \int_{y=b}^{+\infty} m_2([x,y]) dy dx  \\
+\displaystyle m_2([a,b])\int_{x=-\infty}^a \int_{y=b}^{+\infty} m_1([x,y]) dy dx.
\end{array}
\end{eqnarray}
Ainsi le conflit total, étant la masse attribuée sur l'ensemble vide, est donné par~:

\begin{eqnarray}
\label{conj_cont_empty}
\begin{array}{l}
m_\conj(\emptyset)=\displaystyle \int_{z=-\infty}^{+\infty} \left(\int_{x=-\infty}^z \int_{y=z}^{+\infty} m_1([x,z])m_2([z,y]) dy dx \right.\\
~~~~~~~~~~~~~~+ \left. \displaystyle \int_{x=-\infty}^z \int_{y=z}^{+\infty} m_1([z,y]) m_2([x,z]) dy dx\right) dz .
\end{array}
\end{eqnarray}

Cette règle de combinaison reste non idempotente. Ainsi nous pouvons redéfinir l'auto-conflit par l'équation (\ref{autoconf}). 

La $\PCRmo$ peut donc avoir un intérêt, particulièrement dans les cas de conflit important. Elle s'écrit dans le cas de deux experts par~:
\begin{eqnarray}
\label{PCR_cont}
\begin{array}{rcl}
m_\PCR([a,b])&=&m_\conj([a,b]) \\
&+&\frac{m_1([a,b])^2 \displaystyle \int_{x=b}^{+\infty} \int_{y=x}^{+\infty} m_2([x,y]) dy dx}{m_1([a,b])+ \displaystyle \int_{x=b}^{+\infty} \int_{y=x}^{+\infty} m_2([x,y]) dy dx}
+\frac{m_2([a,b])^2 \displaystyle \int_{x=b}^{+\infty} \int_{y=x}^{+\infty} m_1([x,y]) dy dx}{m_2([a,b])+ \displaystyle \int_{x=b}^{+\infty} \int_{y=x}^{+\infty} m_1([x,y]) dy dx}\\
&+&\frac{m_1([a,b])^2 \displaystyle \int_{x=-\infty}^{a} \int_{y=x}^{a} m_2([x,y]) dy dx}{m_1([a,b])+ \displaystyle \int_{x=-\infty}^{a} \int_{y=x}^{a} m_2([x,y]) dy dx}
+\frac{m_2([a,b])^2 \displaystyle \int_{x=-\infty}^{a} \int_{y=x}^{a} m_1([x,y]) dy dx}{m_2([a,b])+ \displaystyle \int_{x=-\infty}^{a} \int_{y=x}^{a} m_1([x,y]) dy dx}
\end{array}
\end{eqnarray}

\subsection{L'espace de définition des fonctions de masse}
Au lieu de considérer l'ensemble de toutes les disjonctions de l'espace de discernement $2^\Theta$, il est possible de considérer l'ensemble de toutes les disjonctions et de toutes les conjonctions de l'espace de discernement, cet ensemble étant noté $D^\Theta$ \cite{Dezert02}. Nous autorisons ainsi des intersections non vide entre deux éléments de l'espace de discernement. Les masses issues des conflits partiels apparaissent donc sur des éléments de $D^\Theta$ et ne sont donc plus réparties sur les éléments de $2^\Theta$.

Selon les applications, il est possible de ne considérer des intersections non nulles seulement pour une partie des éléments de $2^\Theta$. Ainsi, les règles de combinaison précédemment présentées peuvent s'écrire facilement en considérant des classes d'équivalence au lieu des égalités. 

Du fait de la croissance des fonctions de décision (crédibilité, plausiblité et probabilité pignistique) la décision ne peut être prise sur les éléments de $D^\Theta$ correspondant à des intersections. Cependant, certaines applications, comme en traitement des images \cite{Martin06a} où deux classes peuvent se rencontrer sur la zone d'intérêt considérée, peuvent nécessiter la manipulation d'un tel espace. Dans le cadre des fonctions de masses continues, lorsqu'il est possible de se contenter de la combinaison des intervalles de manière discrète, l'intersection des intervalles prend tout son sens et là encore il peut être intéressant de considérer $D^\Theta$.

\section{Applications}
\label{appli}
Nous présentons ici deux applications dans lesquelles la répartition du conflit peut avoir un intérêt. La première est liée à la caractérisation des fonds sous-marin à partir d'images Sonar, la seconde concerne la classification de cibles aériennes à partir de données Radar.

\subsection{Classification des images Sonar}
Les images Sonar sont obtenues à partir des mesures temporelles faites en traînant à l'arrière d'un bateau un Sonar qui peut être latéral, frontal, ou multifaisceaux. Chaque signal émis est réfléchi sur le fond puis reçu sur l'antenne du Sonar avec un décalage et une intensité variable. Pour la reconstruction sous forme d'images un grand nombre de données physiques (géométrie du dispositif, coordonnées du bateau, mouvements du Sonar, ...) est pris en compte, mais ces données sont entachées des bruits de mesures dues à l'instrumentation. A ceci viennent s'ajouter des interférences dues à des trajets multiples du signal (sur le fond ou la surface), à des bruits de chatoiement, ou encore à la faune et à la flore. Les images sont donc entachées d'un grand nombre d'imperfections telles que l'imprécision et l'incertitude.

Il est alors difficile pour un expert humain de caractériser précisément et avec certitude les fonds marins à partir de telles images. Cependant, en milieu sous-marin les capteurs optiques ne permettent pas d'imager rapidement et de grandes surfaces les fonds, ce qui est nécessaire pour de nombreuses applications telles que la navigation de robots autonomes. Il est donc nécessaire de proposer des approches de classifications automatiques des images Sonar. Nous pouvons distinguer 5 types de sédiments~: les roches, le sable, la vase, les cailloutis (zone de petites roches sur du sable ou vase), et les rides (de sable ou de vase, caractérisées par une texture particulière). 

\begin{figure}[htbp]
\begin{center}
\includegraphics[height=5cm]{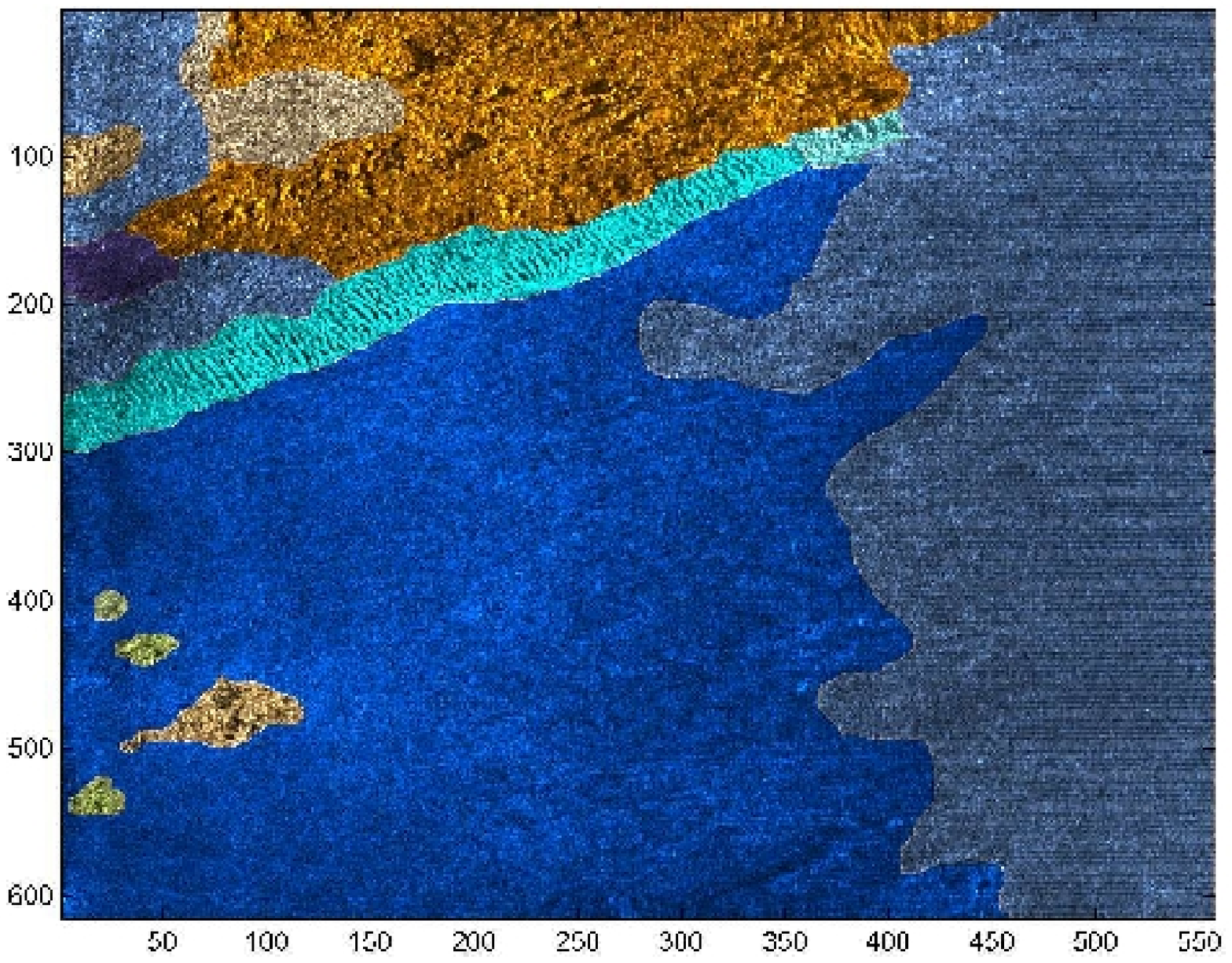}
\includegraphics[height=5cm]{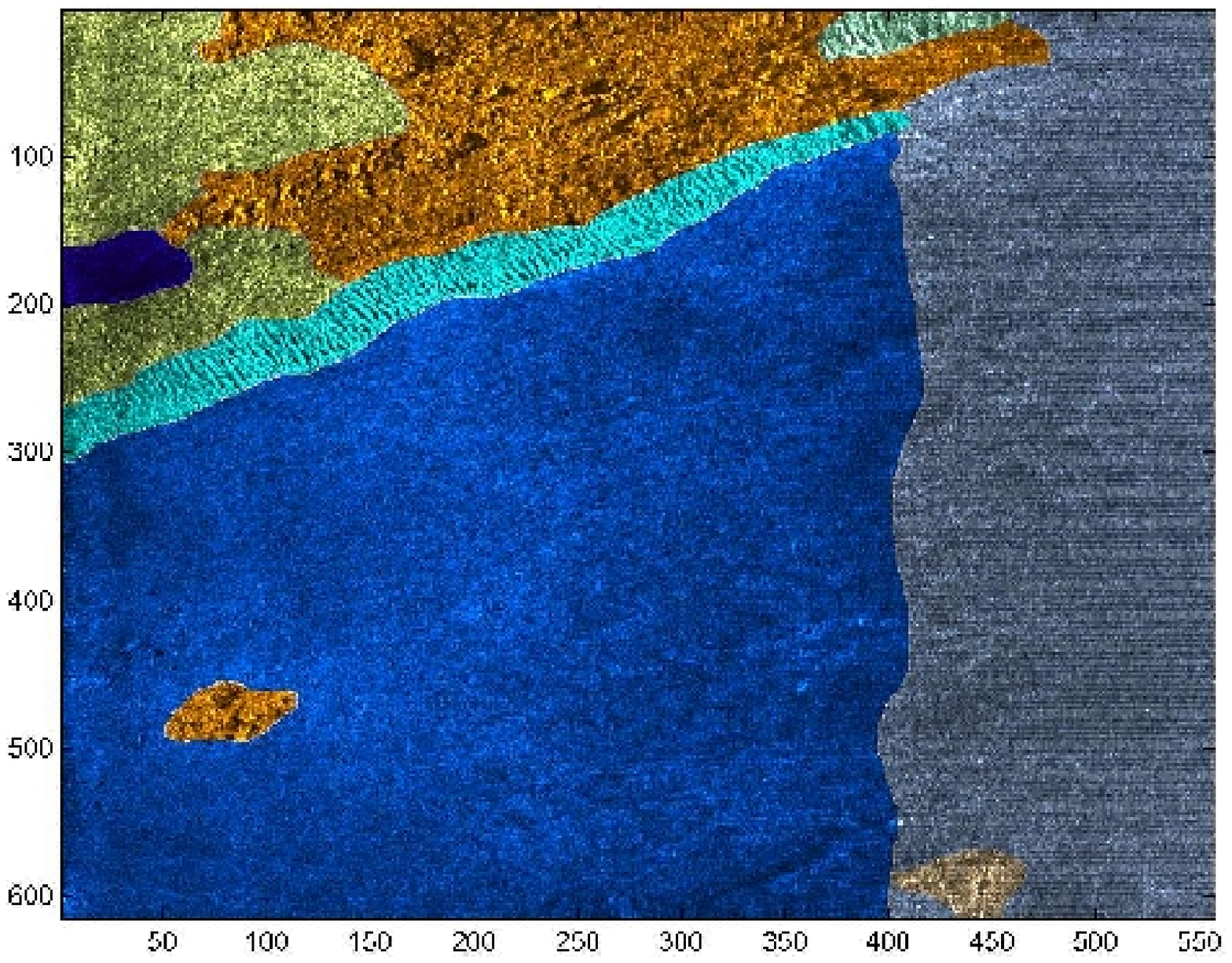}
\caption{Avis de deux experts sur la même image.}
\label{expert}
\end{center}
\end{figure}

La réalité n'étant pas connue précisément, nous avons constitué une base de données à partir de l'avis de différents experts en leur demandant d'indiquer la certitude de l'information fournie (en terme de sûr, moyennement sûr et pas sûr). La figure \ref{expert} présente le résultat de la segmentation obtenue par deux experts, les couleurs indiquant les différents types de sédiments pour une certitude. Nous constatons ainsi le manque de précision de la segmentation et l'incertitude des experts sur leur classification. 

\subsubsection{L'extraction de caractéristiques discriminantes}

La première étape de la classification est l'extraction de paramètres discriminants. Les images Sonar étant texturées, nous allons chercher à caractériser cette texture. La texture d'une image ou d'une partie d'une image est définie par une ressemblance globale qui peut être vue comme la répétition d'un motif plus ou moins grand, pouvant comporter des différences locales très importantes. La texture doit donc être caractérisée au niveau de la globalité de la partie considérée et non au niveau du pixel. Ainsi, des imagettes de taille variable (par exemple 16 $\times$ 16 pixels, ou 32 $\times$ 32 pixels) sont considérées pour la caractérisation de la texture. 

L'extraction de caractéristiques de texture sur des images peut être réalisée à partir de différentes méthodes de décomposition de l'image telle que les matrices de co-occurrence, les décompositions en ondelettes, par exemple à partir des filtres de Gabor, ou encore les longueurs de plage \cite{Martin05}. Une fois l'image décomposée différents paramètres peuvent être calculés tels que énergie, entropie, homogénéité, etc.

Dans cet article nous considérons une décomposition à partir des matrices de co-occurence qui par ailleurs ont donné de bons résultats \cite{Martin05}. Une matrice de co-occurrence est obtenue par une estimation des probabilités de transition de niveaux gris entre deux pixels situés à une distance donnée (ici de un pixel) selon différentes directions $d$~:  0, 45, 90 et 135 degrés. Cette matrice $C_d$ est donc une matrice carrée ayant pour taille le nombre de niveaux de gris considéré $n_g$. A partir de cette décomposition Haralick \cite{Haralick73} a proposé 14 paramètres, nous en avons retenu 6~: l'homogénéité, le contraste, l'entropie, la corrélation, la directivité, et l'uniformité. 

L'\textit{homogénéité} est donnée par~:
\begin{eqnarray}
\sum_{i=1}^{n_g}\sum_{j=1}^{n_g}C_d^2(i,j),
\end{eqnarray}
qui a une valeur élevée pour des images uniformes, ou possédant une texture périodique dans le sens de la translation $d$.

Le \textit{contraste} qui caractérise des probabilités de transition élevées pour des pixels ayant une grande différence de niveau de gris, est donné par~:
\begin{eqnarray}
\frac{1}{n_g-1}\sum_{k=o}^{n_g-1}k^2 \sum_{i,j=1, |i-j|=k}^{n_g}C_d(i,j).
\end{eqnarray}

L'\textit{entropie} qui reste faible s'il y a peu de probabilités de transition élevées dans la matrice, est estimée par~:
\begin{eqnarray}
1-\sum_{i=1}^{n_g}\sum_{j=1}^{n_g}C_d(i,j) \ln(C_d(i,j)).
\end{eqnarray}

La \textit{corrélation} entre les lignes et les colonnes de la matrice est calculée par~:
\begin{eqnarray}
\frac{1}{\sigma_x \sigma_y}\left| \sum_{i=1}^{n_g}\sum_{j=1}^{n_g}(i-m_x)(j-m_y)C_d(i,j)\right|,
\end{eqnarray}
où $m_x$, $m_y$, $\sigma_x$, et $\sigma_y$ sont respectivement les moyennes et écart-types des lignes et de colonnes de la matrice $C_d$.

La \textit{directivité} sera élevée si la texture présente une direction privilégiée dans le sens de la translation $d$~:
\begin{eqnarray}
\sum_{i=1}^{n_g}C_d(i,i).
\end{eqnarray}

L'\textit{uniformité} caractérise la proportion d'un même niveau de gris dans l'image~:
\begin{eqnarray}
\sum_{i=1}^{n_g}C_d^2(i,i).
\end{eqnarray}

Après une étude de ces paramètres sur les images Sonar, nous avons constaté que la taille des imagettes a peu d'influence sur la séparation des histogrammes selon les classes. De même la direction de translation $d$ n'influence pas ou peu la séparation des classes. En effet, seules les rides peuvent avoir des directions privilégiées, cependant nous ne cherchons pas à séparer les rides verticales des rides obliques, mais simplement les rides des autres classes. Ainsi, nous avons choisi de considérer les six paramètres décrits ci-dessus en prenant la moyenne arithmétique sur les quatre directions de $d$. 

\begin{figure}[htbp]
\begin{center}
\includegraphics[height=5cm]{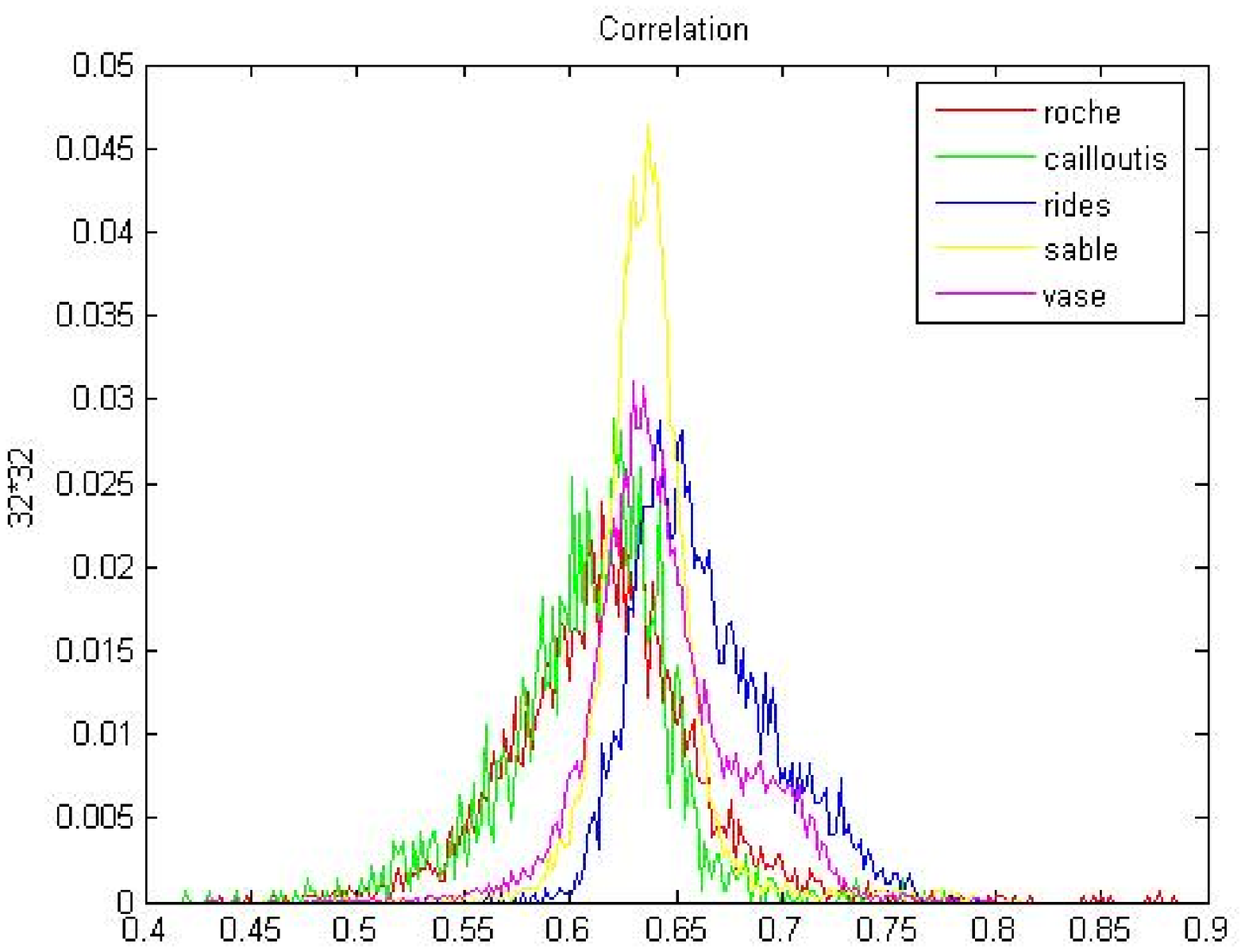}
\includegraphics[height=5cm]{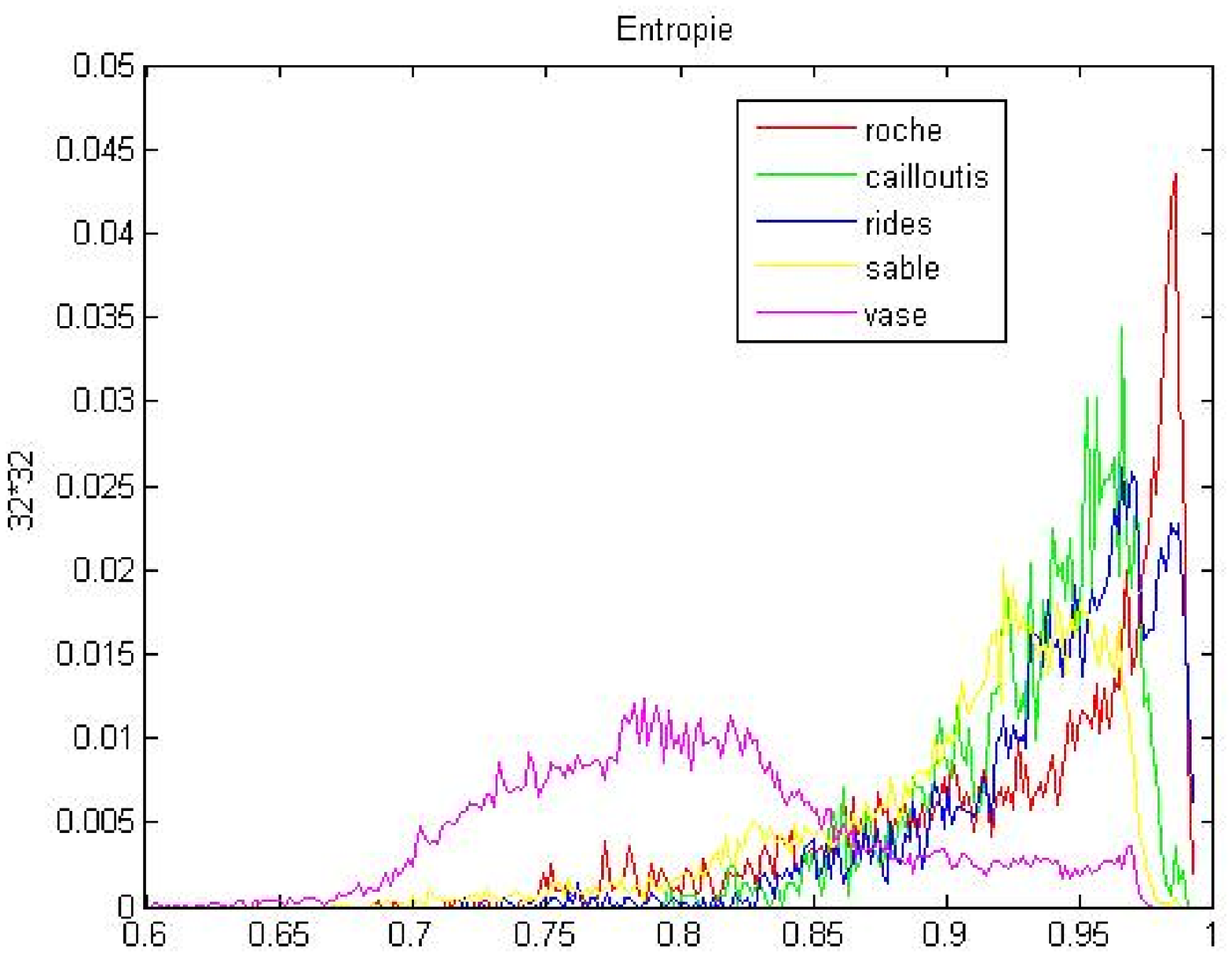}
\caption{Corrélation et entropie.}
\label{corr_ent}
\end{center}
\end{figure}

La figure \ref{corr_ent} montre la difficulté qu'il peut découler de la classification à partir de tels paramètres pour les images Sonar. En effet, nous constatons que pour l'entropie seule la vase se distingue des autres sédiments, alors que pour la corrélation il y a un fort recouvrement. Cette confusion du type de sédiment par un paramètre entraîne un auto-conflit de ce paramètre, ce qui va générer du conflit lors de la combinaison. Ces deux figures résument bien le comportement des quatre autres paramètres.

\subsubsection{Modélisation des fonctions de masse}
Une première approche possible pour les fonctions de masse est celle proposée par Den{\oe}ux \cite{Denoeux95,Zouhal98}~:
\begin{eqnarray}
\label{m_deno}
\left\{
\begin{array}{l}
	m_p(C_i | x^{(t,k)})(x)=\alpha_{ip}\exp\left(\gamma_i d^2(x,x^{(t,k)})\right)\\
	m_p(\Theta | x^{(t,k)})(x)=1-\alpha_{ip}\exp\left(\gamma_i d^2(x,x^{(t,k)})\right)\\
\end{array}
\right.
\end{eqnarray}
où $C_i$ est la classe associée à $x^{(t,k)}$ ({\em i.e.} $d^{(t,k)}=C_i$), et $x^{(t,k)}$ sont les $k$ vecteurs d'apprentissage les plus proches de la valeur $x$ du paramètre $p$. $\alpha_{ip}$ est un coefficient d'affaiblissement, et $gamma_i$ un coefficient de normalisation qui peuvent être optimisés \cite{Zouhal98}.

Les fonctions de masse des paramètres de texture peuvent aussi être définies à partir des histogrammes normalisés obtenus pour un expert. Notons $d_i^p(x)$ le nombre d'imagettes de la classe $C_i$ dont le paramètre de texture $p$ donne $x$. La normalisation proposée est~: $d_i^p(x)/\max_x(\sum_i d_i^p(x))$. Ainsi la masse pour un paramètre $p$ et pour une classe $C_i$ est donnée par~: 

\begin{eqnarray}
\label{m_hist}
\left\{
\begin{array}{l}
m_p(C_i)=\displaystyle \frac{d_i^p(x)}{\max_x(\displaystyle \sum_i d_i^p(x))}\\
m_p(\Theta)=1-\displaystyle \sum_i  m_p(C_i).
\end{array}
\right.
\end{eqnarray}

\subsubsection{Résultats}
Notre base d'images Sonar est composée de 42 images fournies par le GESMA (Groupe d'Etudes Sous-Marines de l'Atlantique). Ces images ont été obtenues à partir d'un Sonar latéral Klein 5400 avec une résolution de 20 à 30 cm en azimut et 3 cm en range dans des profondeurs allant de 15 m à 40 m. Ces images ont été segmentées par différents experts. L'estimation des masses pour chaque paramètre et selon le type de sédiment est effectuée à partir de l'information donnée par un expert, tandis que les tests sont réalisés à partir des informations fournies par un second expert.

Nous ne considérons ici que des imagettes de taille 32$\times$32 pixels ne comportant qu'un seul type de sédiment afin d'obtenir des matrices de confusion classiques. Nous obtenons ainsi une base de 30294 imagettes pour l'expert fournissant l'apprentissage et de 30745 imagettes pour le second expert. Cinq types de sédiments sont considérés~: roche (10\%), cailloutis (5\%), ride (13\%), sable (27\%) et vase (45\%).

Le tableau  \ref{res_sonar} présente les résultats de classification issue de la fusion selon les règles de combinaison orthogonale, de Dubois et Prade et de la $\PCRmo$, des fonctions de masse définies pour chaque paramètre par l'équation (\ref{m_deno}) ou par l'équation (\ref{m_hist}). Il faut prendre garde à ces pourcentages qui peuvent paraître faibles. L'évaluation de la classification des images Sonar est délicate, pour une évaluation plus fine le lecteur peut se reporter à \cite{Martin06d}.

\begin{table}[htbp]
\caption{Pourcentage de bonne classification par type de sédiment.}
\begin{center}
  \begin{tabular}{|l|l|c|c|c|c|c|c|}
    \hline
   & & Roche & Cailloutis & Ride & Sable & Vase & Global\\
    \hline
    & $m_\conj$ & 27.0388 &  18.4974 &  24.4464&   11.3517  & 20.0870 & 18.78 \\
    \cline{2-8}
    équation (\ref{m_deno}) & $m_\DP$ & 16.1290  & 22.8230 &  20.0174  & 21.8504 &  18.4912& 19.72\\
    \cline{2-8}
    & $m_\PCRmo$ & 14.0993  & 19.1235  & 16.0226 &  26.5912 &  25.4023 & 22.64\\
    \hline
   & $m_\conj$ & 23.3323 &   0 & 4.7741  & 79.8283  & 79.7774 & 61.55 \\
    \cline{2-8}
    équation (\ref{m_hist})  & $m_\DP$   & 23.3645 &0  & 4.7741 & 79.8165 & 79.7774 & 61.55 \\
    \cline{2-8}
    & $m_\PCRmo$   &28.6819 & 0 & 8.1273 & 81.0280 & 77.2168 & 61.66 \\
    \hline
  \end{tabular}
\end{center}
\label{res_sonar}
\end{table}

Nous constatons que l'approche proposée par l'équation (\ref{m_deno}) donne de faibles pourcentages en général mais que les résultats pour les classes faiblement représentées sont meilleurs que ceux donnés par l'équation (\ref{m_hist}) (par exemple pour les cailloutis ou les rides). La comparaison des trois approches de combinaison montre que la répartition du conflit apporte de meilleurs résultats qui sont significativement meilleurs même dans le cas de l'équation (\ref{m_hist}) étant donné le grand nombre d'imagettes. Dans une optique d'amélioration des pourcentages il est possible de combiner les masses issues des différentes méthodes de combinaison ou bien de combiner ces classifieurs tel que nous l'avons proposé dans \cite{Martin05}.

\subsection{Classification de cibles Radar}

\section{Conclusions}
Nous avons présenté une nouvelle règle de combinaison en répartissant les conflits partiels sur les éléments le produisant. Les comportements et propriétés de cette règle ont été étudiés en comparaison avec les règles existantes. Cette règle peut être appliquée dans de nombreuses situations lorsque le conflit est important ou non, en discret ou en continue, ou encore dans le cadre de la fusion statique ou dynamique. 
Nous avons présenté les résultats de cette nouvelle combinaison dans le cadre de deux applications~: en imagerie Sonar et en classification de cibles Radar. Les taux de classification montrent que la répartition du conflit apporte de meilleurs résultats que les règles précédemment introduites dans la littérature. Ces résultats ne garantissent pas que les performances de la nouvelle règle de combinaison seront les meilleures dans toutes les applications, mais montrent l'intérêt de celle-ci dans certaines applications.

\bibliographystyle{plain} 
\bibliography{biblio}
\end{document}